\documentclass{article}
\usepackage{PRIMEarxiv}
\usepackage[utf8]{inputenc}
\usepackage[T1]{fontenc}
\usepackage{hyperref}
\usepackage{url}
\usepackage{booktabs}
\usepackage{amsfonts}
\usepackage{nicefrac}
\usepackage{microtype}
\usepackage{fancyhdr}
\usepackage{graphicx}
\usepackage{cite}
\usepackage{amsmath,amssymb,amsfonts}
\usepackage{algorithmic}
\usepackage{textcomp}
\usepackage{multirow}
\usepackage[subrefformat=parens]{subcaption}
\usepackage{caption}
\title{Estimating Driver Personality Traits from On-Road Driving Data}

\author{
  Ryusei Kimura\\
  Japan Advanced Institute of Science Technology \\
  \texttt{s2110059@jaist.ac.jp} \\
   \And
  Takahiro Tanaka \\
  Institute of Innovation for Future Society\\
  Nagoya University \\
  \texttt{tanaka@coi.nagoya-u.ac.jp} \\
    \And
  Yuki Yoshihara \\
  Institute of Innovation for Future Society\\
  Nagoya University \\
  \texttt{y-yuki@coi.nagoya-u.ac.jp} \\
   \And
  Kazuhiro Fujikake \\
  School of Psychology\\
  Chukyo University \\
  \texttt{fujikake@lets.chukyo-u.ac.jp} \\
   \And
  Hitoshi Kanamori \\
  Institute of Innovation for Future Society\\
  Nagoya University \\
  \texttt{hitoshi\_kanamori@coi.nagoya-u.ac.jp} \\
   \And
  Shogo Okada \\
  Japan Advanced Institute of Science Technology \\
  \texttt{okada-s@jaist.ac.jp}
}

\graphicspath{{Figure/}}

\begin{document}
\maketitle

\begin{abstract}
This paper focuses on the estimation of a driver's psychological characteristics using driving data for driving assistance systems.
Driving assistance systems that support drivers by adapting individual psychological characteristics can provide appropriate feedback and prevent traffic accidents.
As a first step toward implementing such adaptive assistance systems, this research aims to develop a model to estimate drivers' psychological characteristics, such as cognitive function, psychological driving style, and workload sensitivity, from on-road driving behavioral data using machine learning and deep learning techniques.
We also investigated the relationship between driving behavior and various cognitive functions, including the Trail Making Test (TMT) and Useful Field of View (UFOV) test, through regression modeling.
The proposed method focuses on road type information and captures various durations of time-series data observed from driving behaviors.
First, we segment the driving time-series data into two road types, namely, arterial roads and intersections, to consider driving situations.
Second, we further segment data into many sequences of various durations.
Third, statistics are calculated from each sequence.
Finally, these statistics are used as input features of machine learning models to estimate psychological characteristics.
The experimental results show that our model can estimate a driver's cognitive function, namely, the TMT~(B) and UFOV test scores, with Pearson correlation coefficients $r$ of 0.579 and 0.708, respectively.
Some characteristics, such as psychological driving style and workload sensitivity, are estimated with high accuracy, but whether various duration segmentation improves accuracy depends on the characteristics, and it is not effective for all characteristics.
Additionally, we reveal important sensor and road types for the estimation of cognitive function.
\end{abstract}


\section{Introduction}
\label{sec:introduction}
The rapid development of the automobile industry has had a significant impact on human life.
While it has brought many benefits to society, the automobile industry has also contributed to an increasing number of problems, including traffic accidents involving older drivers.
In 2020, the Centers for Disease Control and Prevention reported that approximately 7,500 older adults were killed in traffic crashes and that almost 200,000 older adults were treated in emergency departments for crash-related injuries~\cite{centers2021older}.
One of the main causes of traffic accidents by older drivers is cognitive decline.
According to~\cite{owsley1991visual}, older drivers are more prone to accidents and tend to have significant impairments in cognitive function.
However, it is difficult to self-identify cognitive decline and drive according to cognitive ability.

Driving assistance systems represent a promising solution to this problem, but it is not easy to develop driving assistance systems that are acceptable and comfortable for all drivers.
Exiting systems are usually designed based on average driver characteristics~\cite{martinez2017driving} even though there are various types of drivers.
One-size-fits-all driving assistance systems are not suitable for all drivers, and drivers will eventually ignore them.
Therefore, driving assistance systems that support drivers by adapting to individual characteristics are expected to be more helpful for preventing traffic accidents~\cite{bengler2014three}.
For example, personalized driving assistance systems can alert drivers to dangerous locations considering their cognitive function or workload sensitivity.

To develop personalized and acceptable driving assistance systems, automatically monitoring individual characteristics using daily driving data to obtain driver characteristic information is an effective approach.
Automatically estimating these characteristics from driving data can provide useful information for driving assistance systems without requiring burdensome cognitive tests or questionnaires.

Much previous research has focused on driving style recognition from driving data~\cite{martinez2017driving} for personalized driving assistance systems.
However, although psychological driver characteristics such as cognitive ability~\cite{aksan2015cognitive}\cite{Anderson2005Cognitive}, psychological driving style~\cite{ishibashi2007indices}, and workload sensitivity~\cite{ishibashi2008indices}, are related to driving behaviors, researchers have not focused on recognizing these characteristics.
Thus, it is not clear whether the recognition of these driver psychological characteristics is possible.
As a first step toward implementing adaptive assistance systems, the aim of this research is to develop a regression model to estimate a driver's psychological characteristics, such as cognitive function, psychological driving style, and workload sensitivity, from on-road driving behavioral data using machine learning and deep learning techniques.
Several studies have shown the relationship between driving performance and cognitive function~\cite{aksan2015cognitive}\cite{Anderson2005Cognitive}, personality~\cite{adrian2011personality}\cite{classen2011personality}, and stress~\cite{matthews1991personality}\cite{dorn1992two}.
Based on these results, we posit that psychological driver characteristics can be estimated from driving data.

These characteristics are invariant over a short period, such as a single drive; driving data obtained on public roads contain various kinds of driving behaviors observed in diverse situations.
Driving behaviors on public roads vary with road type, driving condition, and situation.
Thus, estimating drivers' psychological characteristics on public roads is a difficult and challenging task, and we do not know in which part of driving data the differences in characteristics will appear.
To address this problem, our proposed method segments driving data based on road types and divides them into small sequences to compare driving behaviors under similar conditions.
We evaluate the proposed method and investigate the effectiveness of the segmentation.
Furthermore, we assessed the use of important in-vehicle sensors for estimation and analyze which characteristics that can be estimated for each road type.

We use the dataset in~\cite{Yoshihara2016Accurate}, which includes time-series driving data collected from in-vehicle sensors, for example, acceleration, brake, and steering angle sensors.
The subjects are older drivers who drive on public roads.
In terms of metrics from driver psychological characteristics, we use the Driving Style Questionnaire (DSQ)~\cite{ishibashi2007indices}, Workload Sensitivity Questionnaire (WSQ)~\cite{ishibashi2008indices}, and several neuropsychology tests, including the Trail Making Test (TMT)~\cite{reitan1958validity}, Maze test~~\cite{ott2008computerized}, and Useful Field of View test (UFOV)~\cite{ball1993useful}.

The main contributions of this paper are as follows.
\begin{enumerate}
\item \textbf{Estimation of drivers' psychological characteristics considering road types:}
The proposed method uses road type information to estimate drivers' psychological characteristics.
Road type information is informative for the estimation of drivers' psychological characteristics because the ability to drive safely differs across different road types~\cite{de2001predicting}.
We introduce the method in Section \ref{Method} and present the evaluation of the performance in Section \ref{Results}.

\item \textbf{Analysis of effective in-vehicle sensors for estimating drivers' psychological characteristics:}
The Effectiveness of in-vehicle sensors for estimating psychological driver characteristics from driving data is still unknown.
We assess the effectiveness of in-vehicle sensors using various types of sensors in Section \ref{Discussion}.

\item \textbf{Analysis of the effective duration of driving behavior for the estimation of drivers' psychological characteristics:}
We reveal how the duration of behavior contains important information about drivers' psychological characteristics in Sections \ref{Results} and \ref{Discussion}.
This knowledge is useful for feature extraction or estimation models that achieve high accuracy.
\end{enumerate}


\section{Related works}
\label{Related works}
This study focuses on estimating drivers' psychological characteristics from driving data.
No previous study has addressed this estimation, but several studies have analyzed the relationship between driving and psychological characteristics.
We hypothesize that psychological characteristics can be estimated from driving data based on previous studies.

\subsection{Relationship between driving and personality}
Some studies have revealed that driving performance is related to personality.
Adrian et al.~\cite{adrian2011personality} investigated the relationship between driving performance and personality traits among older drivers.
It was reported that personality (extraversion) was negatively related to driving performance.
In~\cite{dahlen2012taking}, aggressive driving, crashes, and moving violations were estimated from the driver's personality.
Guo et al.~\cite{guo2016impact} found that drivers' personality traits affect accident involvement and risky driving behavior.
Additionally, an association between driving stress and personality was identified in~\cite{matthews1991personality}.

\subsection{Relationship between driving and cognitive function}
Cognitive function can be measured by several neuropsychology tests, such as the Mini-Mental State Examination (MMSE)~\cite{folstein1975mini}, TMT~\cite{reitan1958validity}, Maze test~\cite{snellgrove2005cognitive}, UFOV~\cite{ball1993useful}.
These scores are used in this study, and details are described in Section \ref{Dataset}.
The relationship between driving and the scores of these tests has been well studied.
Piersma et al.~\cite{piersma2016prediction} assessed fitness to drive in patients with Alzheimer's disease using clinical interviews, neuropsychological assessments, and driving simulator rides.
These three types of assessments are valid for assessing fitness to drive.
Baines et al.~\cite{baines2018meta} showed that the patterns of driving cessation differ depending on the sex of participants with dementia.
The results of these works show that a correlation exists between cognitive function and driving behavior. 

\subsection{Driver characteristics estimation}
Wallace et al.~\cite{wallace2021preliminary} classified drivers as those with Lewy body dementia, Alzheimer's type dementia, and healthy controls.
Grethlein et al.~\cite{grethlein2021adhd}\cite{grethlein2022extracting} estimated drivers with attention deficit hyperactivity disorder (ADHD) from driving simulator data.
These studies were similar to this study, but they used a driving simulation system to collect driving data.
In this study, we use real driving data collected on public roads.
Hence, the experiments are conducted in an environment that is closer to actual conditions.
Wang et al.~\cite{wang2020identifying} used real driving data to estimate a driver's personality.
In~\cite{wang2020identifying}, only driving signals from a straight route were used, while in this study, we use driving data from more road types, such as intersections, to capture diverse driving behavior.

This study significantly expands on the study in~\cite{kimura2022psychological} by adding an estimation of a driver's cognitive function from real driving data.
In~\cite{kimura2022psychological}, only DSQ and WSQ were estimated, and cognitive function was not estimated.
Furthermore, in addition to the proposed model, we use a neural network model (LSTM) that is not used in \cite{kimura2022psychological} to estimate psychological characteristics and verify the performance of the LSTM model.

\begin{table}
\caption{\textbf{In-vehicle sensors.}}
\label{tab:in-vehicle}
\centering
\setlength{\tabcolsep}{3pt}
\begin{tabular}{|c|lr|}
  \hline
  & Sensor & Unit\\ 
  \hline \hline
  1& Steering angle & $deg$\\
  2& EPS torque & $Nm$\\
  3& Forward acceleration & $m/s^2$\\
  4& Lateral acceleration & $m/s^2$\\
  5& Yaw rate & $deg/sec$\\
  6& Speed & $km/h$\\
  7& Accelerator position & $\%$ \\
  8& Brake pressure & $MPa$\\
  9 & Fuel consumption & $ml$\\
  \hline
\end{tabular}
\label{tab1}
\end{table}

\begin{figure*}
  \centering
  \includegraphics[width=13cm]{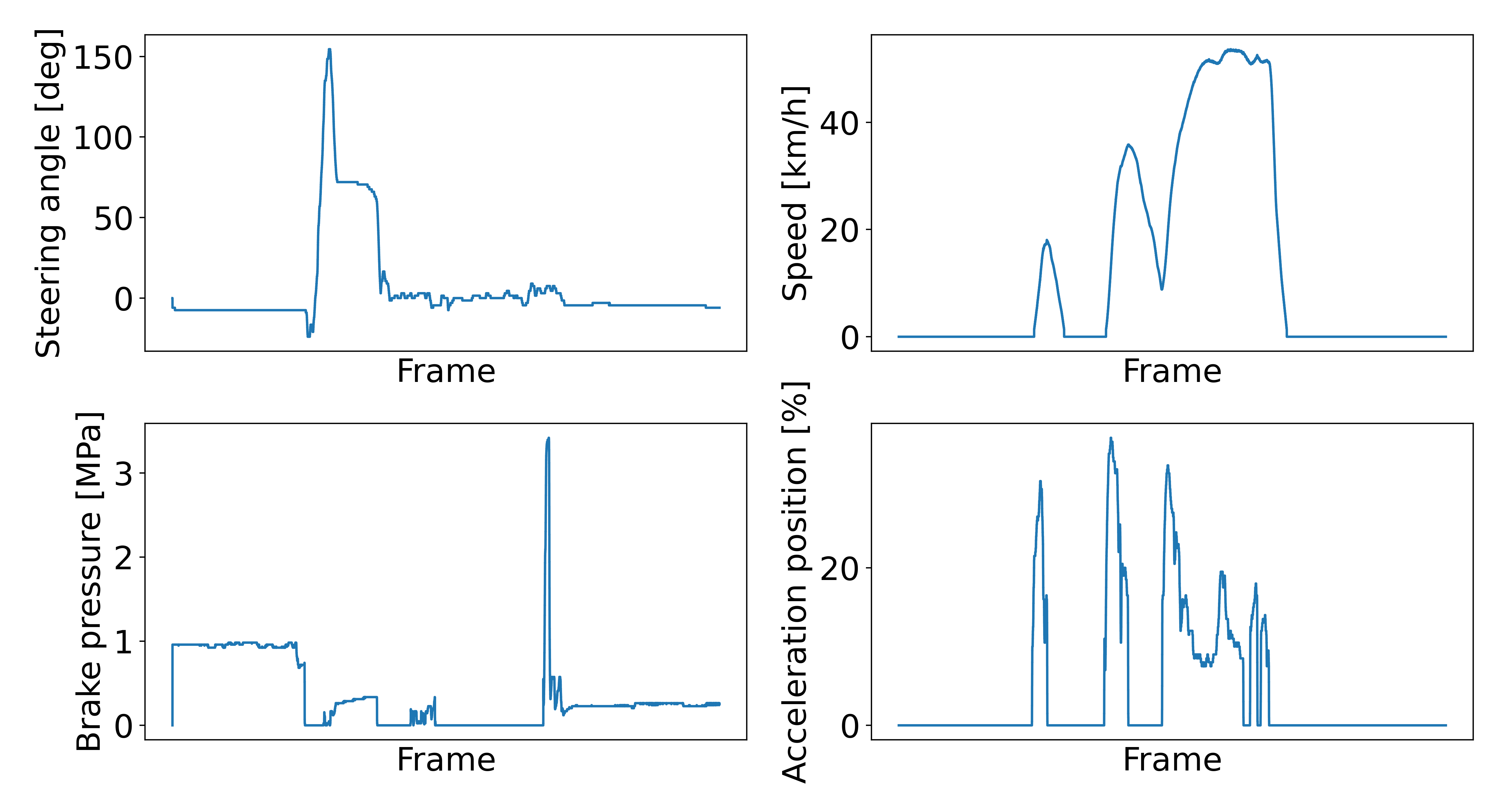}
  \caption{\textbf{Examples of time-series data from in-vehicle sensors.}}
  \label{fig:plot_sisgnals}
\end{figure*}

\section{Dataset}
\label{Dataset}
In this study, we use the dataset provided by the Institutes of Innovation for Future Society of Nagoya University~\cite{Yoshihara2016Accurate}.
The dataset was collected from 24 elderly people (12 males and 12 females) aged 50 to 79 years (average 66 years).
They drove on public roads two times each, and a total of 48 driving session data were collected.
After removing incomplete data, we retained 38 driving session data from 23 drivers.
Hence, for some drivers, one driving session data is utilized.
The Ethical Committee of Nagoya University approved all procedures in this study.
Informed consent was obtained from all drivers before conducting the experiments.

The driving tests were performed on public roads. 
In the tests, all participants drove on an arterial road first and then circumnavigated a residential area.
The total mileage and the total driving time were different for each participant.
The driving duration ranged from 2245 $s$ to 4762 $s$, with an average of 2885 $s$.
The mileage ranged from 10079 $m$ to 14810 $m$, with an average of 12109 $m$.
In addition to the driving tests, the drivers took cognitive function tests and answered questions about their driving style and workload.

\subsection{In-vehicle sensor data}
The car used in the driving tests was equipped with several sensors.
We use time-series data from 9 sensors and GPS sensor data.
The GPS sensor data are used only for preprocessing of driving data.
Table \ref{tab:in-vehicle} details the in-vehicle sensors.
Moreover, we calculate first-order differences in the steering angle, forward acceleration, lateral acceleration, and accelerator position and use them as the velocity of the steering angle, forward jerk, lateral jerk, and rate of change of the accelerator position. 
From these sensors, we estimate the results of cognitive function tests, the DSQ, and the WSQ.
Figure \ref{fig:plot_sisgnals} shows examples of time-series data of the steering angle, speed, brake pressure, and acceleration position sensors.

\begin{table}
\centering
\caption{\textbf{Descriptive statistics, and Pearson correlations among cognitive function tests.}}
\setlength{\tabcolsep}{3pt}
\label{tab:score_corr}
 \begin{tabular}{|c|cccc|cc|}
  \hline
  & TMT~(A) & TMT~(B) & MAZE & UFOV & Mean & SD\\ 
  \hline
  TMT~(A) & $-$ & 0.65 & 0.57 & 0.23 & 34.1 & 10.2\\
  TMT~(B) & & $-$ & 0.51 & 0.53 & 94.9 & 36.3\\
  Maze & & & $-$ & 0.08 & 26.3 & 16.9\\
  UFOV & & & & $-$ & 151.4 & 100.1\\
  \hline
 \end{tabular}
\end{table}

\begin{figure}
  \centering
  \includegraphics[width=12cm]{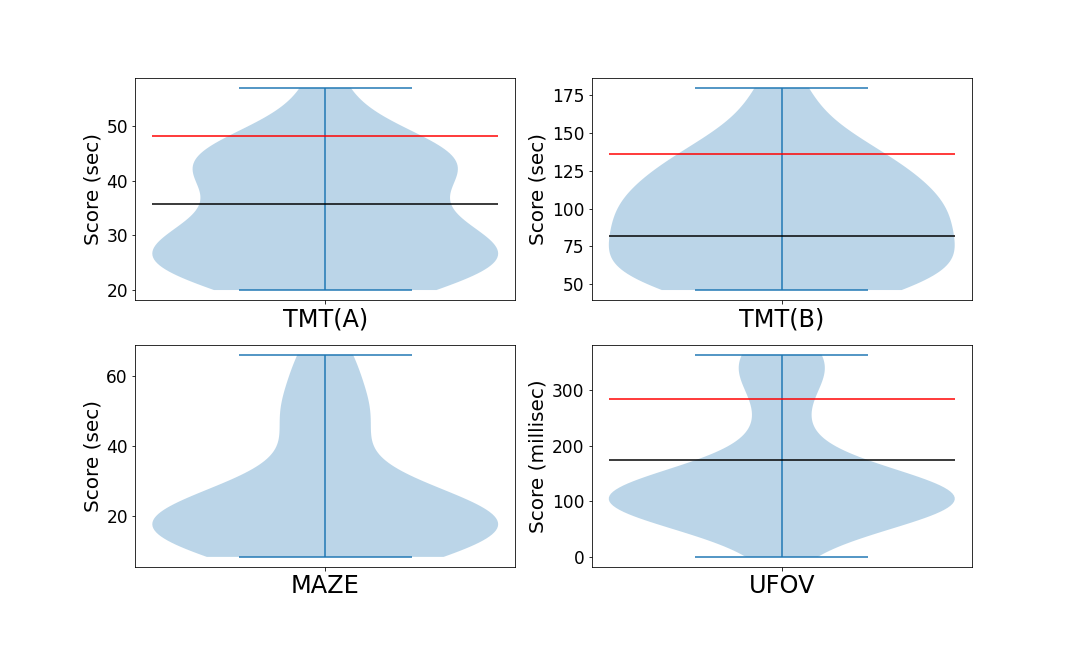}
  \caption{\textbf{Violin plots of the cognitive function scores. The TMT and MAZE scores are expressed in seconds, and the UFOV test scores are expressed in milliseconds. The red and black lines indicate the mild cognitive impairment (MCI) and cognitively normal standards in~\cite{ashendorf2008trail} or passing the driving test in~\cite{samuelsson2021decisions}. The average TMT~(A), TMT~(B), MAZE, and UFOV scores of the subjects were 34.1, 94.9, 26.3, and 151.4, respectively.}}
  \label{fig:violin}
\end{figure}

\subsection{Cognitive function data}
\label{sec:cog}
The dataset also includes scores on neuropsychological tests that measure cognitive function.
The details of the tests used in this study are as follows.

\textbf{Trail Making Test (TMT)}:
There are two types of TMT~\cite{reitan1958validity}.
TMT~(A) measures the execution time of a test in which subjects connect numbers written on a piece of paper in sequence.
In TMT~(B), subjects connect numbers and letters; thus, cognitive alteration is needed.
It has been shown that certain TMT results are correlated with impaired driving in older drivers~\cite{dobbs2013effective} and that these tests are useful as a screening tool~\cite{papandonatos2015clinical}.

\textbf{Maze Test}:
Subjects execute the Maze test~\cite{ott2008computerized}, and the time to finish the test is measured.
In~\cite{ott2008computerized}, the correlation between the results of the Maze test and driving performance was verified.
There are five types of maze tests, and subjects randomly solve two types of tests.
We use the total score from the two types of tests.

\textbf{Useful Field of View (UFOV) test}:
The UFOV test~\cite{ball1993useful} measures the visual field area, where information can be extracted without eye or head movements.
The time to finish the UFOV test is measured.
The results of the UFOV test generally decrease with age~\cite{allison2000effects} and correlate with vehicle accidents~\cite{owsley1991visual}\cite{owsley1998visual}.
Therefore, the UFOV test is considered an important estimator of the behavior of older drivers.

Descriptive statistics and the Pearson correlations among cognitive function tests of 23 drivers are provided in Table \ref{tab:score_corr}.
Some studies investigated the standard scores of cognitive function tests.
Ashendorf et al.~\cite{ashendorf2008trail} investigated the average TMT scores among cognitively normal older adults and adults with MCI.
In~\cite{samuelsson2021decisions}, patients with cognitive dysfunction took a driving test; their UFOV test scores were examined in the passing and failing groups.
Figure \ref{fig:violin} shows violin plots of the test scores.
These scores are plotted in Figure \ref{fig:violin}, from which we can confirm that the scores of each test are widely distributed.

In addition to these tests, drivers took the MMSE~\cite{folstein1975mini}, which is a screening test for dementia estimation.
In this test, subjects answer some questions.
Subjects who score less than 23 out of 30 are suspected to have dementia, and those who score less than 28 are suspected to have MCI.
All drivers included in this dataset had MMSE scores in the range of 28 to 30.
Thus, no driver was suspected of having dementia. The MMSE score is not incorporated in the subsequent analysis.

\subsection{Driving Style Questionnaire and Workload Sensitivity Questionnaire}
The DSQ and WSQ, which are based on a self-reported questionnaire, were introduced by~\cite{ishibashi2007indices}\cite{ishibashi2008indices} for characterizing drivers from a psychological aspect.
In~\cite{ishibashi2008indices}, the relationship between car-following behavior and DSQ was validated.
Table \ref{tab:DSQandWSQ} details the items in the DSQ and WSQ.
The DSQ includes eight items measured on a scale from 1 to 4, and the WSQ includes 10 items measured on a scale from 1 to 5.
We classify drivers with high and low scores on the DSQ and WSQ.

\begin{table}[t!]
\centering
\caption{\textbf{DSQ and WSQ items.}}
\label{tab:DSQandWSQ}
\setlength{\tabcolsep}{3pt}
 \begin{tabular}{|c|l|}
  \hline
  & DSQ Item\\ 
  \hline \hline
  1& Confidence in driving skill\\
  2& Hesitation for driving\\
  3& Impatience in driving\\
  4& Methodical driving\\
  5& Preparatory maneuvers at traffic signals\\
  6& Importance of automobile for self-expression\\
  7& Moodiness in driving\\
  8& Anxiety about traffic accidents\\
  \hline
  \hline
  & WSQ Item\\ 
  \hline \hline
  1& Understanding of traffic conditions\\
  2& Understanding of road conditions\\
  3& Interference with concentration \\
  4& Decline in physical activity\\
  5& Disturbance on the pace of driving\\
  6& Physical pain\\
  7& Path understanding and search\\
  8& In-vehicle environment\\
  9& Control operation\\
  10& Driving posture\\
  \hline
 \end{tabular}
\end{table}

\begin{figure*}[t]
  \begin{minipage}[t]{0.48\linewidth}
    \centering
    \includegraphics[width=7.8cm]{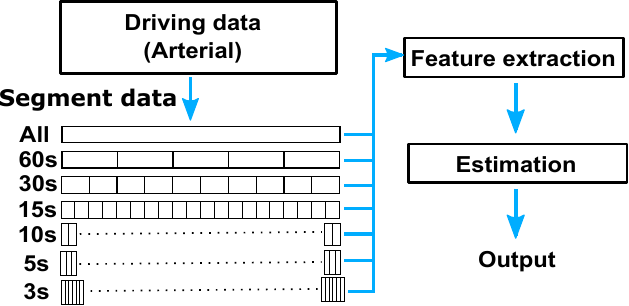}
    \subcaption{\textbf{Overview of the proposed model in arterial roads.}}
  \end{minipage}
  \begin{minipage}[t]{0.48\linewidth}
    \centering
    \includegraphics[width=7.8cm]{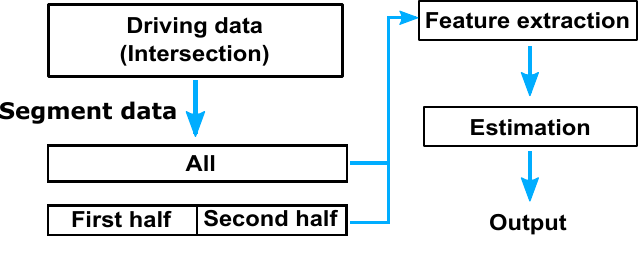}
    \subcaption{\textbf{Overview of the proposed model in intersections.}}
  \end{minipage}
    \caption{\textbf{Overview of the proposed model.}}
  \label{fig:overview}
\end{figure*}

\begin{figure}
    \centering
    \includegraphics[width=8cm]{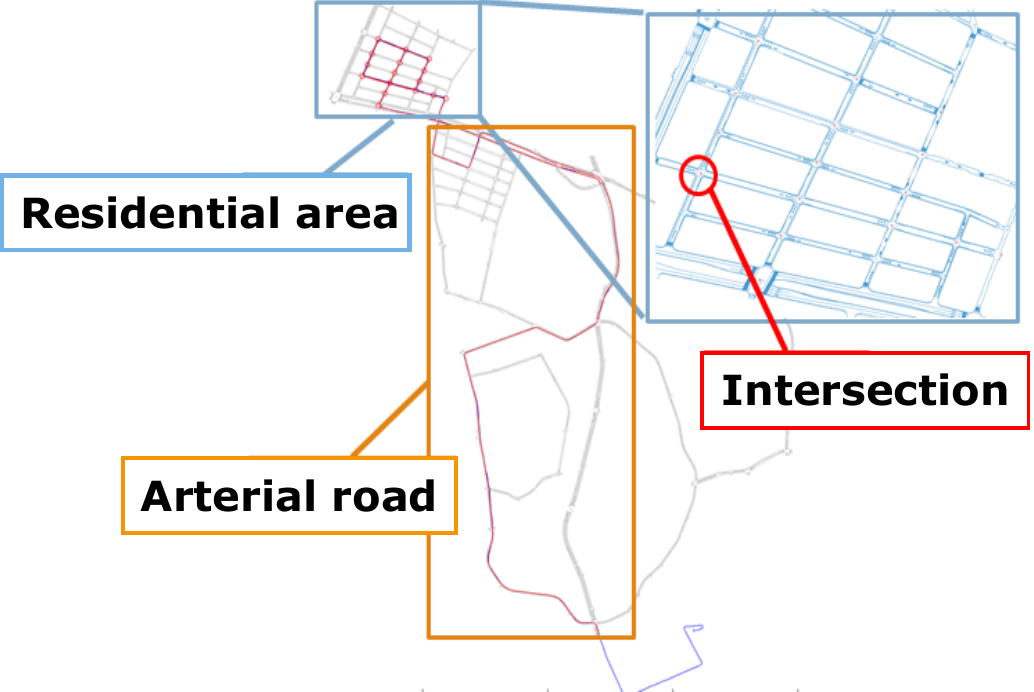}
    \caption{\textbf{Driving test route (red line). Time-series driving data are segmented into arterial roads and intersections. The intersections are located in a residential area.}}
    \label{fig:route}
\end{figure}

\begin{figure}
    \centering
    \includegraphics[width=8.5cm]{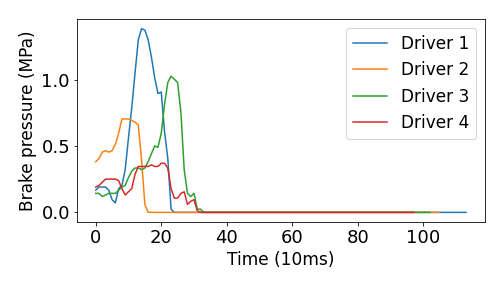}
    \caption{\textbf{Examples of the brake pressure behavior of four drivers. The drivers press the brake pedal when entering an intersection.}}
    \label{fig:brake}
\end{figure}

\section{Method}
\label{Method}
In this section, we present the method for estimating driver characteristics from in-vehicle sensors.
An overview of the proposed method is shown in Figure \ref{fig:overview}.
We make two hypotheses for the estimation as follows (in Section \ref{HP1} and \ref{HP2}).

\subsection{Segmentation using road types}
\label{HP1}
First, we hypothesize that driver characteristics, which can be estimated are dependent on the road type.
Since driving behavior is strongly related to road type, several studies on driving style recognition from driving data have used road type information~\cite{murphey2009driver}\cite{aguilar2017different}.
In addition, older drivers tend to be involved in traffic accidents at intersections~\cite{mcgwin1999characteristics}.
Additionally, the mental workload in driving depends on road context~\cite{bongiorno2017driver} and visibility~\cite{baldwin2004mental}.
Thus, it is natural that features calculated from different road type data elicit different driver characteristics.
Moreover, segmenting driving time-series data based on road types makes capturing diverse driving behaviors easier.
Based on this hypothesis, we segment time-series driving data into two types, namely, arterial roads and intersections, to consider driving situations.
The detailed method of road type segmentation is described in Section \ref{Preprocessing}.

\subsection{Segmentation with various durations}
\label{HP2}
Second, we hypothesize that differences in drivers' characteristics are exhibited not only in whole driving but also in partial driving.
However, we have no a priori knowledge about when or where the differences in drivers' characteristics are exhibited.
Therefore, we segment arterial road and intersection data further with various durations and extract features from each segmented road.

We segment arterial data so that the average number of seconds for each segment is [All, 60, 30, 15, 10, 5, 3].
"All" means no division, i.e., whole arterial road (with an average of 355 $s$).
On the other hand, at intersections, it is difficult to segment driving data based on the average number of seconds, such as arterial road data, because the number of seconds to pass through the intersection is small compared with the arterial road data and varies according to the situation.
Thus, we segment the data at the intersection into the first half and the second half.

Long-term driving behavior is captured from features from long-term driving data (segment), while short-term driving behavior is captured from features from short-term driving data (segment).
This second hypothesis is similar to the hypothesis in~\cite{grethlein2021adhd} in that only a few subintervals of time-series driving data are indicative of the class of drivers with ADHD.
The detailed method of various durations segmentation is described in Section \ref{Preprocessing}.

\subsection{Preprocessing}
\label{Preprocessing}
To segment driving data using road types, we use the car's position obtained through a GPS sensor.
The route of the driving test is shown in Figure \ref{fig:route}.
The intersections are located in a residential area.
We use four intersections where all drivers' data are recorded accurately.
We treat these four intersections as the same road types, but features are extracted from each intersection separately because the visibility and ease of driving are not the same.
The time-series driving data collected on the arterial road and at selected intersections are segmented into arterial road data and intersection data.

Furthermore, we segment time-series data on the arterial road.
We position segment points so that the average number of seconds for each segment is [ALL (355), 60, 30, 15, 10, 5, 3].
Then, each frame of time-series data is assigned to the nearest segment points, and the arterial road data are segmented into several intervals.

For intersection data, we use brake sensor data to segment the data.
When entering an intersection, almost all drivers press the brake pedal, step off the brake, and then step on the accelerator.
Figure \ref{fig:brake} shows the braking behaviors of some drivers at an intersection.
Since it is assumed that driving behavior changes before and after stepping off the brake, we segment intersection data into before and after stepping off the brake.
We found that small parts of the driving behaviors of some drivers at intersections do not follow this pattern.
We exclude these driving data and extract intersection features (the sample size used in the experiment remains the same).

\subsection{Feature extraction}
\label{Feature_extraction}
For each road type and segment in Section \ref{Method}, statistics (mean, median, variance, maximum, kurtosis, and skewness) of each sensor in Table \ref{tab:in-vehicle} are calculated and used as features for estimation models.
In total, the numbers of features are 17051 for the arterial road and 905 for the intersection data (missing features are removed). 
The differences in the features for the arterial road and the intersections are the location and intervals, where driving data are collected.

We estimate cognitive function, DSQ, and WSQ using driving sensors.
Regression is performed for cognitive function, and classification is performed for DSQ and WSQ because the scores of cognitive function are continuous values and the scores of DSQ and WSQ are discrete values.
The input features are the same for estimation models of cognitive function, DSQ, and WSQ.

\subsection{Machine learning algorithm}
For both regression and classification, we separately estimate the characteristics of drivers on arterial roads and at intersections and compare the estimation results based on two types of roads.
To estimate the scores of cognitive function tests, we use linear regression models: lasso regression, ridge regression, and nonlinear regression models: random forest and long short-term memory (LSTM).
To estimate the DSQ and WSQ scores, we use linear classification models, i.e., logistic regression with L2 regularization and linear support vector, and nonlinear classification models: random forest and LSTM.
We use an LSTM model as a deep learning method that can capture complex time-series relationships between the sensors and cognitive function.
An LSTM model is also used in~\cite{grethlein2022extracting} to estimate drivers with ADHD.

The models other than LSTM use the statistical features extracted in Section~\ref{Feature_extraction}.
The estimation models based on arterial road data use 17051 features, and the estimation models based on intersection data use 905 features.
In contrast, LSTM uses the time series signal features of sensors in Table~\ref{tab:in-vehicle}.
The input signals of the LSTM are sampled at 1 Hz.
LSTM is used to test the first hypothesis in Section~\ref{HP1}, but it cannot be used for the second hypothesis explicitly because LSTM uses only time-series signals.
From intersection data, the LSTM model makes estimations for each intersection because time-series data are directly input, and features extracted from several intersections cannot be utilized.
All models output estimated scores of each cognitive function test or class categories of DSQ and WSQ.

\section{Experimental settings}
\label{Experimental setting}

\subsection{Experimental settings for cognitive function estimation}
\label{Experimental setting cognitive}
The regularization parameter values of lasso and ridge regression are selected from [0.001, 0.01, 0.1, 1, 10, 100].
The maximum depth of a tree of random forest is selected from [3, 5, 7, 9, 11].
To align the lengths of all time-series signals, the first part of the time-series data is filled with zeros.
The LSTM architecture is composed of one hidden layer with 50 units.
The mean squared error (MSE) loss is used as the loss function for the LSTM.
The number of epochs is 5000 in all models, but the learning is terminated when the value of the loss function does not decrease 10 times in a row.
Optimization is performed using the ADAM optimizer with a learning rate of 0.001.
As evaluation criteria of estimation models for cognitive function, the Pearson correlation coefficient ($r$) and root-mean-square-error (RMSE) are used.

\subsection{Experimental settings for DSQ and WSQ estimation}
The regularization parameter values of the logistic regression and linear support vector machine are selected from [0.001, 0.01, 0.1, 1, 10, 100].
The maximum depth of a tree of the random forest is selected from [3, 5, 7, 9, 11].
The architecture of the LSTM models is almost the same as that of regression in Section \ref{Experimental setting cognitive}, but the binary cross entropy loss is used as the loss function.
Model training proceeds in the same manner as regression.
As an evaluation criterion of classification models, we report the macro F1-score.
The scales of DSQ and WSQ are different; we split these scores based on the median value to create binary classification labels and then, conduct binary classification.

For both regression and classification, we use leave-one-person-out cross-validation to evaluate the estimation models.
For drivers whose two driving data are used, two estimated scores are aggregated using mean values and are regarded as the final estimated score of drivers to balance a driver's contribution to accuracy.
For models other than LSTM, for each fold, we use only features that have a correlation with true scores with $|r|>0.1$ to avoid overfitting.
The hyperparameters are tuned in the training set. 

\section{Results}
\label{Results}
\begin{table*}
\centering
\caption{\textbf{Regression accuracy of the model with two types of segmentation.}}
\label{tab:res_cog}
\scalebox{1.0}{
 \begin{tabular}{|c|c|cccc|cccc|}
 \hline
  \multirow{2}{*}{Road type} & \multirow{2}{*}{Model} & \multicolumn{4}{c|}{$r$} & \multicolumn{4}{c|}{RMSE}\\
  \cline{3-10}
   &  & TMT~(A) & TMT~(B) & MAZE  & UFOV & TMT~(A) & TMT~(B) & MAZE & UFOV\\
  \hline
  \multirow{3}{*}{Arterial} & Lasso & 0.213 & 0.532 & \textbf{0.334} & -0.150 & 10.963 & 31.794 & \textbf{15.939} & 146.331\\
  & Ridge & \textbf{0.416}* & \textbf{0.579}* & 0.281 & -0.149 & \textbf{9.477} & \textbf{29.234} & 16.279 & 130.413\\
  & RF & 0.102 & -0.403 & -0.164 & -0.189 & 10.282 & 39.618 & 17.389 & 106.429\\
  \hline
  \multirow{3}{*}{Intersection} & Lasso & -0.340 & -0.008 & -0.282 & 0.080 & 12.072 & 44.755 & 20.535 & 121.543\\
  & Ridge & -0.188 & 0.066 & -0.415 & 0.186 & 14.813 & 47.959 & 28.415 & 114.376\\
  & RF & -0.298 & -0.077 & -0.391 & \textbf{0.559}* & 11.560 & 39.049 & 19.061 & \textbf{86.287}\\
  \hline
 \end{tabular}}
\end{table*}

\begin{table*}
\centering
\caption{\textbf{Regression accuracy of the model without various duration segmentation.}}
\label{tab:res_dur_cog}
\scalebox{1.0}{
 \begin{tabular}{|c|c|cccc|cccc|}
 \hline
  \multirow{2}{*}{Road type} & \multirow{2}{*}{Model} & \multicolumn{4}{c|}{$r$} & \multicolumn{4}{c|}{RMSE}\\
  \cline{3-10}
   &  & TMT~(A) & TMT~(B) & MAZE  & UFOV & TMT~(A) & TMT~(B) & MAZE & UFOV\\
  \hline
  \multirow{3}{*}{Arterial} & Lasso & 0.073 & 0.313 & 0.244 & 0.055 & 10.691 & 35.603 & 16.178 & 107.376\\
  & Ridge & 0.400 & 0.224 & \textbf{0.411}* & -0.134 & 9.838 & 37.893 & \textbf{15.085} & 123.433\\
  & RF & 0.207 & -0.062 & 0.068 & -0.111 & 10.226 & 38.816 & 17.420 & 108.093\\
  \hline
  \multirow{3}{*}{Intersections} & Lasso & -0.353 & 0.214 & -0.120 & 0.439* & 11.606 & 36.655 & 18.526 & 90.759\\
  & Ridge & 0.213 & 0.284 & -0.025 & 0.316 & 10.804 & 38.722 & 20.639 & 114.131\\
  & RF & 0.128 & 0.055 & -0.273 & \textbf{0.708}* & 10.307 & 38.026 & 18.794 & \textbf{81.564}\\
  \hline
 \end{tabular}}
\end{table*}

\begin{table*}
\centering
\caption{\textbf{Regression accuracy of the model without any segmentation.}}
\label{tab:res_road_cog}
\scalebox{1.0}{
 \begin{tabular}{|c|cccc|cccc|}
 \hline
  \multirow{2}{*}{Model} & \multicolumn{4}{c|}{$r$} & \multicolumn{4}{c|}{RMSE}\\
  \cline{2-9}
   & TMT~(A) & TMT~(B) & MAZE  & UFOV & TMT~(A) & TMT~(B) & MAZE & UFOV\\
  \hline
  Lasso & -0.226 & 0.205 & -0.124 & -0.345 & 12.237 & 37.883 & 19.314 & 129.162\\
  Ridge & 0.279 & 0.403 & 0.094 & -0.320 & 10.497 & 33.721 & 18.342 & 150.510\\
  RF & -0.087 & 0.362 & 0.091 & 0.426 & 11.688 & 33.784 & 17.101 & 91.490\\
  \hline
 \end{tabular}}
\end{table*}

In this section, we evaluate the proposed model and confirm the efficacy of two types of segmentation.
We compare the estimation accuracies of three models, namely, (i) a model with both road type and various duration segmentation, (ii) a model with only road type segmentation, and (iii) a model without any segmentation.
The features used in model (ii) are extracted from each road type without various duration segmentation.
The features used in model (iii) are extracted from whole driving data without any segmentation.
The accuracy of the LSTM model is low for cognitive function, DSQ, and WSQ.
The estimation results of the LSTM are detailed in Appendix \ref{App:1}.

\subsection{Regression accuracy for cognitive function}
Table \ref{tab:res_cog} shows the regression accuracy of model (i) for the arterial road and intersection data.
The first column shows the road types that are used for feature extraction.
Lasso regression, ridge regression, and random forest are denoted as Lasso, Ridge, and RF, respectively, in the second column.
The bold values indicate the highest accuracy among each cognitive function test.
We performed t-tests to validate the null hypothesis of no correlation between estimated values and labels.
Correlation coefficients with p-values smaller than 0.05 are marked with * in Table \ref{tab:res_cog}.

Using ridge regression for the arterial road data, TMT~(A) and TMT~(B) were estimated, with $r$ values of 0.416 and 0.579, respectively.
These $r$ values indicate a moderate correlation between estimated scores and true scores, namely, the estimation worked well for TMT~(A) and TMT~(B).
However, the accuracies for these tests were not high in the intersection models.
MAZE was best estimated with lasso regression for the arterial road, and the $r$ value was 0.334.
Using a random forest for the intersection data, UFOV had an $r$ value of 0.559.
In~\cite{mukaka2012guide}, correlations between 0.1 and 0.5 are interpreted as weak correlations, those between 0.5 and 0.7 as moderate correlations, and those between 0.7 and 0.9 as strong correlations.
Thus, the highest estimation accuracies of model (i) for TMT~(A) and MAZE were weak correlations, and those for TMT~(B) and UFOV were moderate.
The estimation models that achieved the best RMSE for each cognitive function test were not always the same as the estimation models that achieved the best $r$.
The best RMSE values of all tests were smaller than the standard deviation (in Figure \ref{tab:score_corr}).
Hence, we find that the estimation by our models was better than the estimation by mean values.

In Table \ref{tab:res_cog}, the estimations for TMT~(A), TMT~(B), and MAZE were better for the arterial roads than those at the intersections.
On the other hand, the results of UFOV were worse for the arterial roads.
These results confirm that cognitive functions, which can be estimated depend on the road type.
As~\cite{owsley1991visual} reported that older drivers with a useful-field-of-view disorder had 15 times more intersection accidents, and it is assumed that the difference in the UFOV results of drivers is likely to be apparent at an intersection.

All computations were carried out on an Intel Core i7-12700K CPU with 12 cores and 32 GB RAM.
Preprocessing and feature extraction took a total of 55.7 seconds per driver for the arterial road data, and 40.0 seconds per driver for the intersection data.
We also measured the time spent during CV to train and test the models.
Ridge regression for the arterial road was the best model to estimate TMT~(B), and it took an average of 1.33 seconds to learn the driving data of 22 drivers and an average of 0.02 seconds to estimate the driving data of one driver.

\subsubsection*{Effectiveness of various duration segmentation}
Next, to confirm the effectiveness of various duration segmentation, we compare the accuracies of the models with road type and various duration segmentation (i) and the models only with road type segmentation (ii). 
Table \ref{tab:res_dur_cog} shows the regression accuracies of model (ii).
The bold values indicate the accuracies that were higher than the best accuracies of model (i).

For model (ii), the accuracies of MAZE and UFOV were high in comparison with model (i).
For intersections in particular, improvements in UFOV were significant.
Model (ii) achieved a moderate $r$ value in MAZE (0.411) and a strong $r$ value in UFOV (0.708).
These results illustrate that the various duration segmentation worked well for TMT~(A) and TMT~(B) on arterial roads but did not work well for the intersections.

\subsubsection*{Effectiveness of road type segmentation}
As described in Section \ref{Method}, we hypothesized that road types are informative for the estimation of the driver's cognitive function.
To test this hypothesis, we compare the accuracies of the models with road type segmentation (i) and (ii) and the models without road type segmentation (iii).
Table \ref{tab:res_road_cog} shows the regression accuracies of model (iii).
None of the tests were estimated more accurately than models with road type segmentation.
Therefore, it is important to consider the road type when estimating the cognitive functions of drivers.

\subsection{Classification accuracy for DSQ and WSQ}
\begin{table*}
\centering
\caption{\textbf{Classification accuracy (F1-score) for DSQ.}}
\label{tab:res_DSQ}
\setlength{\tabcolsep}{3pt}
\scalebox{1.0}{
 \begin{tabular}{|c|ccc|ccc||ccc|}
 \hline
 \multirow{2}{*}{DSQ item} & \multicolumn{3}{|c|}{Model (i)} & \multicolumn{3}{|c||}{Model (ii)} & \multicolumn{3}{|c|}{Model (iii)}\\
  & LR & SVM & RF & LR & SVM & RF & LR & SVM & RF\\
 \hline
 \hline
 \multicolumn{7}{|c||}{Arterial} & \multicolumn{3}{c|}{Whole}\\
 \hline
 Confidence in driving skill & 0.356 & 0.387 & 0.415 & 0.415 & 0.360 & 0.406 & 0.415 & 0.315 & 0.387\\
 Hesitation for driving & 0.321 & 0.296 & 0.387 & 0.387 & 0.512 & 0.345 & 0.296 & \textbf{0.649} & 0.432\\
 Impatience in driving & 0.397 & 0.397 & 0.415 & 0.460 & 0.389 & 0.406 & 0.415 & 0.333 & 0.367\\
 Methodical driving & 0.356 & 0.377 & 0.387 & 0.360 & 0.374 & 0.367 & 0.424 & 0.415 & 0.360\\
 Preparatory maneuvers at traffic signals & 0.308 & 0.378 & 0.378 & 0.526 & 0.472 & 0.482 & 0.568 & 0.603 & 0.526\\
 Importance of automobile for self-expression & 0.333 & 0.333 & 0.333 & 0.255 & 0.415 & 0.374 & 0.470 & 0.550 & 0.506\\
 Moodiness in driving & 0.507 & 0.432 & 0.356 & 0.552 & 0.491 & 0.544 & 0.404 & 0.491 & 0.525\\
 Anxiety about traffic accidents & 0.424 & 0.424 & 0.424 & 0.507 & 0.513 & 0.475 & 0.562 & \textbf{0.683} & 0.424\\
 \hline
 \hline
 \multicolumn{7}{|c||}{Intersection}\\
 \cline{1-7}
 Confidence in driving skill & 0.377 & 0.309 & 0.397 & 0.406 & 0.408 & 0.387\\
 Hesitation for driving & 0.616 & 0.591 & 0.418 & 0.389 & 0.391 & 0.356\\
 Impatience in driving & 0.367 & 0.404 & 0.397 & 0.432 & \textbf{0.525} & 0.397\\
 Methodical driving & 0.321 & 0.359 & 0.356 & 0.345 & 0.434 & 0.432\\
 Preparatory maneuvers at traffic signals & 0.591 & 0.670 & 0.470 & 0.755 & \textbf{0.784} & 0.582\\
 Importance of automobile for self-expression & 0.582 & \textbf{0.661} & 0.424 & 0.397 & 0.482 & 0.415\\
 Moodiness in driving & \textbf{0.553} & 0.512 & 0.474 & 0.513 & 0.491 & 0.457\\
 Anxiety about traffic accidents & 0.415 & 0.406 & 0.424 & 0.424 & 0.415 & 0.424\\
 \cline{1-7}
 \end{tabular}}
\end{table*}
\begin{table*}
\centering
\caption{\textbf{Classification accuracy (F1-score) for WSQ.}}
\label{tab:res_WSQ}
\scalebox{1.0}{
 \begin{tabular}{|c|ccc|ccc||ccc|}
 \hline
 \multirow{2}{*}{WSQ item} & \multicolumn{3}{|c|}{Model (i)} & \multicolumn{3}{|c||}{Model (ii)} & \multicolumn{3}{|c|}{Model (iii)}\\
  & LR & SVM & RF & LR & SVM & RF & LR & SVM & RF\\
 \hline
 \hline
 \multicolumn{7}{|c||}{Arterial} & \multicolumn{3}{c|}{Whole}\\
 \hline
 Understanding of traffic conditions & 0.449 & 0.513 & 0.360 & 0.330 & 0.232 & 0.454 & 0.391 & 0.352 & 0.454\\
 Understanding of road conditions& 0.548 & 0.493 & 0.345 & 0.491 & 0.482 & 0.468 & 0.548 & 0.574 & \textbf{0.627}\\
 Interference with concentration & 0.232 & 0.265 & 0.269 & 0.375 & 0.444 & 0.438 & 0.394 & \textbf{0.550} & 0.298\\
 Decline in physical activity & 0.277 & 0.331 & 0.352 & 0.282 & 0.391 & 0.308 & 0.191 & 0.421 & 0.338\\
 Disturbance on the pace of driving & 0.472 & \textbf{0.514} & 0.434 & 0.265 & 0.391 & 0.406 & 0.359 & 0.391 & 0.384\\
 Physical pain & 0.255 & 0.284 & 0.247 & 0.415 & \textbf{0.681} & 0.338 & 0.391 & 0.444 & 0.361\\
 Path understanding and search & 0.247 & 0.170 & 0.103 & 0.472 & 0.500 & 0.289 & 0.419 & 0.521 & 0.394\\
 In-vehicle environment & 0.223 & 0.339 & 0.330 & 0.526 & 0.591 & 0.391 & 0.622 & 0.598 & 0.373\\
 Control operation & \textbf{0.578} & 0.491 & 0.391 & 0.368 & 0.419 & 0.500 & 0.472 & 0.419 & 0.472\\
 Driving posture & 0.404 & 0.309 & 0.367 & 0.356 & 0.415 & 0.309 & 0.321 & 0.454 & 0.345\\
 \hline
 \hline
 \multicolumn{7}{|c||}{Intersection}\\
 \cline{1-7}
 Understanding of traffic conditions & 0.512 & \textbf{0.582} & 0.391 & 0.435 & 0.460 & 0.391\\
 Understanding of road conditions & 0.545 & 0.574 & 0.438 & 0.591 & 0.574 & 0.545\\
 Interference with concentration & 0.482 & \textbf{0.550} & 0.394 & 0.428 & \textbf{0.550} & 0.308\\
 Decline in physical activity & 0.579 & \textbf{0.631} & 0.521 & 0.525 & 0.552 & 0.521\\
 Disturbance on the pace of driving & 0.339 & 0.406 & 0.449 & 0.454 & 0.438 & 0.449\\
 Physical pain & 0.265 & 0.438 & 0.223 & 0.359 & 0.391 & 0.373\\
 Path understanding and search & 0.579 & 0.552 & 0.419 & \textbf{0.632} & 0.578 & 0.550\\
 In-vehicle environment & \textbf{0.725} & 0.683 & 0.491 & 0.622 & 0.709 & 0.460\\
 Control operation & 0.521 & 0.438 & 0.521 & 0.552 & 0.474 & 0.521\\
 Driving posture & 0.321 & 0.449 & 0.321 & 0.309 & 0.491 & 0.333\\
 \cline{1-7}
 \end{tabular}}
\end{table*}

Tables \ref{tab:res_DSQ} and \ref{tab:res_WSQ} show the classification results of DSQ and WSQ of model (i), model(ii), and model (iii), respectively.
LR, SVM, and RF denote logistic regression, support vector machine, and random forest.
The bold values indicate the highest accuracy among each item with values exceeding 0.5.

The DSQ items with the highest accuracy greater than 0.5 were hesitation for driving, impatience in driving, preparatory maneuvers at traffic signals, importance of automobile for self-expression, moodiness in driving, and anxiety about traffic accidents.
In particular, the macro F1-score of preparatory maneuvers at traffic signals was high (0.784) at intersections.
The highest accuracy of two out of eight DSQ items did not exceed 0.5 with all models.
Other than driving posture, all WSQ items were estimated with the highest accuracy greater than 0.5.
In particular, the macro F1-score of in-vehicle environment was high (0.725) at intersections.
Concerning DSQ, none of the models using arterial data achieved the highest accuracy using model (i) or model (ii).
Concerning WSQ, the model using arterial data achieved the highest accuracy on three items, while the model using intersection data achieved the highest accuracy on six items.

Model (i) achieved the highest accuracies for two DSQ items and six WSQ items, while model (ii) achieved the highest accuracies for two DSQ items and three WSQ items.
Thus, segmentation with various durations worked well for some items of the psychological characteristics but did not work for other items.
For both DSQ and WSQ, estimation at intersections tends to be more accurate than on arterial roads.
With model (iii), two DSQ items and two WSQ items are estimated with the highest accuracy.
Based on these results, for some DSQ and WSQ items, segmentation based on road type was not effective for estimation.

\section{Discussion}
\label{Discussion}
In this section, we further investigated the efficacy of two types of segmentation.
We focused on the results of the estimation of cognitive function and analyzed why the segmentation improved the accuracies using feature importance.

\begin{table}
    \centering
    \caption{\textbf{The three most important sensors for TMT~(B) and UFOV estimation.}}
    \begin{tabular}{|l|cc|}
    \hline
    \multicolumn{3}{|c|}{TMT~(B) (arterial)}\\
    \hline
    1 & Rate of change of the accelerator position & 75.0\% \\ 
    2 & Steering angle & 21.7\% \\
    3 & Lateral jerk & 1.7\%\\
    \hline
    \hline
    \multicolumn{3}{|c|}{UFOV (Intersection)}\\
    \hline
    1 & Rate of change of the accelerator position & 84.6\%\\
    2 & Accelerator position & 9.81\%\\
    3 & Steering angle & 3.11\%\\
    \hline
    \end{tabular}
    \label{tab:coef_sensor}
\end{table}

\begin{table}
    \centering
    \caption{\textbf{Relative proportion of importance of each segment duration.}}
    \setlength{\tabcolsep}{3pt}
    \begin{tabular}{|c|c|c|}
    \hline
    Duration of segment & Unnormalized & Normalized\\
    \hline
    \hline
        All & 0.1\% & 1.9\%\\
        60$s$ & 9.3\% & 30.3\%\\
        30$s$ & 18.1\% & 29.7\%\\
        15$s$ & 17.6\% & 16.2\%\\
        10$s$ & 19.4\% & 12.2\%\\
        5$s$ & 17.0\% & 5.7\%\\
        3$s$ & 18.6\% & 3.9\%\\
    \hline
    \end{tabular}
    \label{tab:coef_lengths}
\end{table}

\subsection{Contribution of sensors}
We investigate which sensors contributed to the estimation.
We focus on the results of TMT~(B) with model (i) using ridge regression for the arterial roads and those of UFOV with model (ii) using random forest for the intersections.
We regard the absolute values of the standardized regression coefficient of ridge and the mean of accumulation of the impurity decrease within each tree of random forest as the contribution of each feature.
The absolute values are summed for each sensor.
Then, we compare the relative proportions of the three most important sensors.

Table \ref{tab:coef_sensor} shows the relative proportions of the three most important sensors.
For the arterial roads, the three most important sensors and their relative proportions were the rate of change of the accelerator position (75.0\%), steering angle (21.7\%), and lateral jerk (1.7\%), while for the intersections, the three most important sensors and their relative proportions were the rate of change of the accelerator position (84.6\%), accelerator position (9.81\%), and steering angle (3.11\%).
The rate of change of the accelerator position and steering angle are commonly important sensors for both arterial roads and intersections.
At intersections, the accelerator position was a unique effective sensor, and furthermore, the rate of change of the accelerator position was more important.
These important sensor differences between road types were also observed in other tests.
Thus, these differences were caused by the differences among road types rather than those among cognitive function tests.
Moreover, we assume that these differences occurred because driving behavior depends on the road type or driving scene and road type segmentation could capture these differences.

\subsection{Contribution of segmentation}
We analyze which segments worked well for the estimation in the same way as in the previous subsection.
We focus on the results of TMT~(B) of model (i) using ridge regression for the arterial roads and then aggregate feature importance for each segment duration.

Table \ref{tab:coef_lengths} shows the relative proportions of the importance of each duration of the segment.
We compare duration importance which was normalized by the number of segments or not normalized because the number of segments was different depending on the duration of segments.
The left side of Table \ref{tab:coef_lengths} shows the relative proportion without normalization, and the right side shows that with normalization.
The unnormalized importance of segments [All, 60, 30, 15, 10, 5,] were [0.1\%, 9.3\%, 18.1\%, 17.6\%, 19.4\%, 17.0\%, 18.6\%], respectively, while the normalized importance of segments [All, 60, 30, 15, 10, 5,] were [1.9\%, 30.3\%, 29.7\%, 16.2\%, 12.2\%, 5.7\%, 3.9\%], respectively.
Unnormalized proportions demonstrate that every duration, except for "All" of the segment, contributed to the estimation to some extent.
On the other hand, the normalized proportion implies that many segments with short durations were not used much.
This is because important driving behaviors appear not in whole driving but in partial driving.
Moreover, segments with a short duration are noisier than segments with a long duration.

\section{Conclusion}
\label{Conclusion}
In this paper, we addressed a challenging task, estimating the psychological characteristics of drivers, such as cognitive function, psychological driving style, and workload sensitivity, from on-road driving data. Our proposed model uses two types of segmentation, namely, road type segmentation and various duration segmentation, to capture driving behavior.

For cognitive function items, capturing road type information and various durations of driving behavior made the estimation accuracy high.
The best $r$ values of the TMT and UFOV tests were 0.579 and 0.708, respectively.
The experimental results demonstrated that considering the road type improved the accuracy of estimation.
Both long-term and short-term driving behaviors contributed to the estimation on arterial roads and improved the regression accuracy.
Additionally, we confirmed that the rate of change of the accelerator position and steering angle were effective for estimation on arterial roads and at intersections.
For psychological driving style and workload sensitivity, the two types of segmentation improved the accuracy for some items, but they were less effective compared to cognitive function and their effectiveness depended on the items.
This study provides a baseline estimation of psychological characteristics from driving data and benefit analysis.

Our proposed method can be used in situations where all drivers drive the same route because driving data segmentation requires GPS data.
In the actual operating environment, all drivers follow different routes, and the proposed method cannot be applied in such a situation.
Therefore, a future research direction is to focus on particular driving behaviors such as lane changing, overtaking, and curves without GPS data, and then, extract effective features.

This study used driving sensors to estimate the driver's characteristics automatically, but other data sources may contribute to the estimation.
Physiological signals, such as heart rate and skin conductance may reflect the driver's characteristics.
Additionally, camera sensors can be used to capture drivers' behaviors, for example, eye movements and head direction.
The above sensor information is easily available and imposes little burden on the driver.

\bibliographystyle{IEEEtran}
\bibliography{Access.bib}

\section*{Additional experimental results}
\label{App:1}

Here, we detail the estimation results of the LSTM model.
Table \ref{tab:lstm_cog} shows the regression accuracy of the LSTM model for cognitive function.
The LSTM model did not achieve the best accuracy for all cognitive function items.
It is thought that the LSTM model could not learn well because the sample size was small.
Table \ref{tab:lstm_DSQandWSQ} shows the classification accuracy of the LSTM model for DSQ and WSQ.
Similar to the result of the estimation of cognitive function, the LSTM model did not work well for all items and none of the LSTM models achieved an accuracy greater than 0.5.
The accuracy of an LSTM model is expected to improve when the amount of training data increases.
However, it is difficult to collect considerable driving data and characteristic data.
Thus, we need to develop a method that can capture a relationship between driving signals and driver characteristics with few data.

\begin{table*}[h]
\centering
\caption{\textbf{Regression accuracy of the LSTM model for cognitive function.}}
\label{tab:lstm_cog}
\scalebox{0.95}{
 \begin{tabular}{|c|c|cccc|cccc|}
 \hline
  \multirow{2}{*}{Model} & \multirow{2}{*}{Road type} & \multicolumn{4}{c|}{$r$} & \multicolumn{4}{c|}{RMSE}\\
  \cline{3-10}
   & & TMT~(A) & TMT~(B) & MAZE  & UFOV & TMT~(A) & TMT~(B) & MAZE & UFOV\\
  \hline
  \hline
  LSTM & \multirow{1}{*}{Arterial} & 0.216 & 0.252 & -0.084 & -0.022 & 11.278 & 35.903 & 18.419 & 101.603\\
  \hline
  LSTM (Int 1) & \multirow{4}{*}{Intersection} & -0.222 & -0.156 & -0.084 & 0.113 & 11.490 & 34.252 & 16.035 & 97.078\\
  LSTM (Int 2) & & -0.066 & 0.109 & 0.026 & -0.110 & 12.726 & 35.297 & 16.653 & 104.196\\
  LSTM (Int 3) & & -0.469 & -0.243 & -0.013 & -0.157 & 14.191 & 35.581 & 16.115 & 93.642\\
  LSTM (Int 4) & & -0.031 & -0.054 & -0.012 & 0.183 & 11.850 & 34.834 & 16.404 & 95.538\\
  \hline
 \end{tabular}}
\end{table*}

\begin{table*}[ht]
\centering
\caption{\textbf{Classification accuracy of the LSTM model for DSQ and WSQ.}}
\label{tab:lstm_DSQandWSQ}
\scalebox{0.9}{
 \begin{tabular}{|c|c|cccc|}
 \hline
 & \multicolumn{1}{|c|}{Arterial} & \multicolumn{4}{|c|}{Intersection}\\
 \hline
 \hline
 DSQ item & LSTM & LSTM (Int 1) & LSTM (Int 2) & LSTM (Int 3) & LSTM (Int 4)\\
 \hline
 Confidence in driving skill & 0.333 & 0.337 & 0.352 & 0.340 & 0.354\\
 Hesitation for driving & 0.387 & 0.387 & 0.401 & 0.404 & 0.399\\
 Impatience in driving & 0.415 & 0.418 & 0.425 & 0.424 & 0.424\\
 Methodical driving & 0.396 & 0.392 & 0.399 & 0.388 & 0.402\\
 Preparatory maneuvers at traffic signals & 0.355 & 0.359 & 0.369 & 0.356 & 0.371\\
 Importance of automobile for self-expression & 0.377 & 0.374 & 0.368 & 0.357 & 0.363\\
 Moodiness in driving & 0.387 & 0.392 & 0.391 & 0.394 & 0.398\\
 Anxiety about traffic accidents & 0.424 & 0.411 & 0.416 & 0.423 & 0.409\\
 \hline
 \hline
 WSQ item & LSTM & LSTM (Int 1) & LSTM (Int 2) & LSTM (Int 3) & LSTM (Int 4)\\
 \hline
 Understanding of traffic conditions & 0.366 & 0.370 & 0.375 & 0.342 & 0.364\\
 Understanding of road conditions& 0.355 & 0.356 & 0.367 & 0.381 & 0.356\\
 Interference with concentration & 0.355 & 0.338 & 0.357 & 0.361 & 0.340\\
 Decline in physical activity  & 0.344 & 0.353 & 0.374 & 0.375 & 0.358\\
 Disturbance on the pace of driving & 0.333 & 0.345 & 0.368 & 0.368 & 0.350\\
 Physical pain & 0.355 & 0.362 & 0.382 & 0.383 & 0.365\\
 Path understanding and search & 0.333 & 0.342 & 0.364 & 0.366 & 0.347\\
 In-vehicle environment & 0.355 & 0.360 & 0.381 & 0.378 & 0.365\\
 Control operation & 0.333 & 0.354 & 0.350 & 0.349 & 0.360\\
 Driving posture & 0.387 & 0.366 & 0.387 & 0.385 & 0.373\\
 \hline
 \end{tabular}}
\end{table*}

\end{document}